\documentclass[12pt,english,twoside]{article}

\usepackage[a4paper, margin=1in]{geometry}
\usepackage[hang,flushmargin]{footmisc}
\usepackage{setspace}
\usepackage{titlesec}
\usepackage{lipsum}

\renewcommand{\setstretch{1.4}}
\titleformat{\section}[hang]
{\normalfont\bfseries}
{\thesection.}{0.5em}{}
\titleformat{\subsection}[hang]
{\normalfont\bfseries}
{\thesubsection.}{0.5em}{}
\titleformat{\subsubsection}[hang]
{\normalfont\bfseries}
{\thesubsubsection.}{0.5em}{}

\usepackage[colorlinks=true, linkcolor=blue, citecolor=blue, urlcolor=blue]{hyperref} %Hyperref of Citations, delete for submission

\usepackage{amsmath,amssymb} 
\usepackage{tabularray}

\usepackage{tabularx}

\usepackage{ragged2e}
\newcolumntype{P}[1]{>{\RaggedRight\arraybackslash}p{#1}}

\usepackage{xcolor}
\usepackage{makecell}
\usepackage{tabularray}
         
\usepackage{booktabs}
\usepackage{authblk}

\usepackage{csquotes}

\usepackage[style=apa]{biblatex} % 'backend=biber' is the default
\newcommand{\uproman}[1]{\uppercase\expandafter{\romannumeral#1}}

\usepackage{comment}

\usepackage{chngcntr}
\counterwithin{figure}{section}

\title{How Complexity Contributes to Learning Opacity in Machine Learning\textsuperscript{$\dagger$}}
\author[1,2]{Joachim Stein}
\author[2,1]{Eric Raidl}
\affil[1]{Heidelberger Akademie der Wissenschaften}
\affil[2]{Universität Tübingen, Cluster of Excellence: Machine Learning -- New Perspectives for Science}

\begin{document}
\date{\vspace{-5ex}}
\maketitle

\begingroup
\let\thefootnote\relax % Fußnotenmarke temporär deaktivieren
\footnotetext{$\dagger$ This work has been funded by the Deutsche Forschungsgemeinschaft (EXC number 2064/1, project no. 390727645) and the WIN program of the Heidelberg Academy of Sciences and Humanities, financed by the Ministry of Science, Research and the Arts of the State of Baden-Württemberg. For helpful discussion and feedback we thank Sarah Bopp, Timo Freiesleben, Miriam Klopotek, Jan-Willem Romeijn and his group, Hans Rott, Tom Sterkenburg, Max Weinmann, Sebastian Zezulka and the participants of the SAS25-Uncertainty conference in Stuttgart.}
\endgroup

\begin{abstract}
Machine learning (ML) algorithms are known to be opaque. We do not know the reasons for their predictions. The learning process leading to the prediction function is also opaque. We do not fully understand the time evolution of the weight values of neural nets (NN) and related dynamical phenomena.  While prediction opacity is widely studied, learning opacity remains largely underexplored. This article studies learning opacity trough the lens of complex dynamical systems. We argue that NN learning is essentially a complex system and that learning opacity is due to dynamical complexity and the epistemological challenges that arise from it. We identify three key properties of training complexity -- sensitivity to weight initialization, feedback in gradient based optimization, and sensitivity to the training data -- and show how each contributes to learning opacity. As these properties are fundamental to the learning process damping or eliminating  them would fundamentally alter how ML systems learn. 
Some sources of opacity in ML may hence be irreducible.
\end{abstract}

\section{Introduction}\label{sec:Introduction}
%\textcolor{red}{Delete some citations. For example Zuchowski complexity.} 
%\par 
%\textcolor{red}{Footnote to physical systems that we are aware. Maybe at open system.}
%\par
%\textcolor{red}{Terms in one quotation mark `term'}
%\par 
%\textcolor{red}{Normal quotes also with one quotation marks}
%\par 
%\textcolor{red}{Go over indices and weight definitio (Wolfgang confusion)}
%\par 
%\textcolor{red}{Consider, do not, don't}
%\par 
Machine learning, in particular neural networks (NNs), is a powerful automation of induction and learning in general. Given a vast amount of data, machine learning algorithms are able to learn a reliable prediction function.\footnote{In machine learning research, the learned prediction function is often referred to as `the model'. However, to avoid confusion with scientific models used to represent systems, we will use the term \textit{prediction function}.} However, the learned prediction function as well as the learning mechanism themselves remain relatively incomprehensible to us \textemdash \ they are termed `opaque' \parencite{Boge2022, Burrell, Sogaard_2023, Creel_2020, Sullivan_2022}.
\par 
On the one hand we do not know what real-world patterns are encoded in the prediction function. Call this \textit{prediction opacity}. Prediction opacity raises epistemic and ethical challenges in contexts like policy making \parencite{Vredenburgh-2024} or scientific research \parencite{Miriam_Review_Interpretable}. The need to overcome prediction opacity has lead to the field of `eXplainable Artifical Intelligence' (XAI) \parencite[cf.][]{bordt2025position, molnar2025, DearXAICommu}. Researchers in this field develop interpretable machine learning 
methods and provide post-hoc explanation techniques for established machine learning methods. Existing philosophical contributions to XAI investigate, among other things, the reason for prediction opacity \parencite{Sullivan_ML_understanding}, analyse ways to counteract it \parencite{Creel_2020} and how using opaque NN changes the way we gain understanding in science \parencite{Boge2022}. 
\par 
On the other hand, we do not understand \textit{how} the NN has learned. We take this to mean that we do not understand the dynamics of the learning process and related phenomena. Call this \textit{learning opacity}. Learning opacity is also a troublesome phenomenon for even the designers of NNs, who have full access to the learning process, might not understand how the network learns. This is reflected in the great effort of NN researchers to explain phenomena of NN training like benign-overfitting \parencite{belkin2021fitfearremarkablemathematical}, implicit regularization \parencite{neyshabur2015searchrealinductivebias, Falsification_Buchholz_Raidl},  saturation \parencite{understanding_Glorot}, grokking \parencite{power2022grokkinggeneralizationoverfittingsmall} or optimization at the edge of stability \parencite[]{cohen2021gradient}. These phenomena are surprising giving the orthodox view on optimization, and understanding them would help researchers gain better insights into the peculiarities of the optimization process in NNs. Furthermore, since the prediction function is a direct product of the learning process, illuminating learning opacity most likely also advances the analysis of prediction opacity.
\par 
Understanding the sources of learning opacity is also a necessary condition to develop means to counteract it, allowing researchers to deepen their understanding of the NN's learning process. This would have practical implications for debugging, improving training efficiency and real-world deployment. However, unlike in the case of prediction opacity, a thorough conceptual analysis of the sources for learning opacity is still lacking.
\par  
In this article, we provide this conceptual analysis of learning opacity in NNs using insights from complexity science. That even full access to the learning mechanism does not resolve learning opacity shows that it is not merely a matter of limited access. Rather, learning opacity might be due to more intrinsic properties of the optimization heuristics \parencite[cf.][]{Burrell, Creel_2020,Sullivan_2022}. We explore this idea in more depth. We clarify the nature of learning opacity in NN and how complexity may be a major contributing factor to it. Drawing on complexity analysis from the sciences, we identify three characteristic features of complex dynamics \textemdash \ sensitive dependence on initial conditions, feedback and context sensitivity \textemdash \ and show how each limits understanding of systems in the natural sciences. We demonstrate that these properties are also present in NN training, and examine how each of them contributes to learning opacity. 
\par 
The article is structured as follows. Section \ref{sec:MachineLearning} outlines the NN learning process and discusses learning opacity. Section \ref{sec3:Complexity} introduces the notion of complexity from the natural sciences, argues that it provides a conceptual lens for investigating learning opacity, and identifies three central properties of complex systems: sensitive dependence on initial conditions, feedback, and context sensitivity. It also introduces dynamical modelling, a central heuristic for understanding dynamical processes. The remainder of the paper examines these three properties, how they limit researchers’ understanding of natural complex systems, and how they contribute to learning opacity in machine learning. Section \ref{sec:SI} focuses on sensitive dependence on initial conditions, Section \ref{sec:Feedback} on feedback, and Section \ref{sec:CS} on context sensitivity.

\section{Neural Networks}\label{sec:MachineLearning}

Machine learning (ML) refers to the process of using a computer algorithm to learn patterns from data. In this process, researchers provide data to a learning algorithm that aims to identify a function that encodes these patterns and is thus capable of predicting future observations, the \textit{prediction function}.\footnote{For an introduction to the theory of ML, see \textcite{Shalev-Shwartz_Ben-David_2014}. \textcite{Goodfellow-et-al-2016} is the standard textbook on NN learning. \textcite{ Sterkenburg_NFL} and \textcite{Sterkenburg2025} provide a philosophical discussion of the foundations of ML.} In this work, we focus on supervised ML. 
\par 
In supervised ML, a \textit{data point} $(x^{(i)}, y^{(i)}) \in \mathcal{X} \times \mathcal{Y}$ consisting of an instance $x^{(i)}$ from some \textit{instance space} $\mathcal{X}$, e.g. a picture of an animal, paired with its label $y^{(i)}$ from the \textit{label space} $\mathcal{Y}$, e.g. the name of the animal. The process is termed supervised because the learning algorithm has access to the instance's true label during training.
\par 
The prediction function is selected from a predetermined class of functions $f: \mathcal{X} \rightarrow \mathcal{Y}$, known as the \textit{hypothesis class}. In the case of NN learning, the hypothesis class consists of NNs. A NN is a mathematical function that can be represented as a graph $(V, E)$, where $V$ denotes the set of nodes, the \textit{neurons}, and $E$ the set of edges transmitting information from the output of one neuron to the input of another. This work focuses on feed-forward networks, in which the graph is directed and acyclic, so that all edges point in the same direction and no cycles occur. Accordingly, a feed-forward NN is organized as a layered graph.
\par 
A neuron's outputted information is transformed by the neuron's \textit{activation function} $\sigma$, which acts as a threshold function. This transformed information is further \textit{weighted} by a factor $w_i: E \rightarrow \mathbb{R}$ that determines how much information is transmitted along a given edge from one neuron to the next. The vector $w$ consists of all weight factors $w_i$.\footnote{In this paper we use $w_i$ to denote a weight variable or its specific value. Accordingly, $w$ may denote vector containing variables or scalars. The intended meaning of $w_i$ and $w$ should be clear from the context.} A neuron is  defined as the function obtained by applying the activation function to the weighted sum of its input information plus a constant term, the \textit{bias}.\footnote{To simplify the discussion in this paper, we ignore bias terms in the following and focus solely on the weights of the NN.} A specific neuron value is the output of this function for a given input. A NN, $f_w$, is defined as the \textit{repeated composition of neurons}. The number of neurons, their distribution within the network, and the activation function are specified by the \textit{architecture} of the network. The architecture defines a set of neural networks \{$f_w$\} that share the same structural properties but differ in their specific weight and bias values. This set constitutes the hypothesis class, that is, the space of possible prediction functions that may be selected during training. A concrete NN from this set is identified by specifying its weight parameters.
The architecture is a \textit{hyperparameter}. In general hyperparameters are parameters that are needed to define a NN and the learning process, but are neither the weights nor the biases. Hyperparameters for the learning process are, for example, the \textit{learning rate}, the \textit{batch size} or the \textit{weight initialization strategy}. 
\par 
The predictive success of a NN is quantified by the \textit{loss} function: $\mathcal{L}: \{f_w\} \times (\mathcal{X} \times \mathcal{Y}) \rightarrow \mathbb{R}$.\footnote{Common loss functions are the 0-1 loss, $\mathcal{L}[f_{w}(x^{(i)}),y^{(i)}] =1$ if $f_{w}(x^{(i)}) \not= y^{(i)}$ and 0 otherwise and the squared loss $\mathcal{L}[f_{w}(x^{(i)}),y^{(i)}] = (y^{(i)} - f_{w}(x^{(i)}))^2$.\label{fn:activation functions}}  
The goal of the learning process is to find a NN that reliably predicts the true labels for (unseen) input instances from some instance space. To achieve this goal the NN is trained on a finite subset of the whole instance space, the \textit{training data}. The training process searches for the NN with the lowest \textit{empirical risk}, that is, the network whose predictions deviate least, on average, from the true labels of the training instances. The deviation is measured by the chosen loss function. To identify the NN that minimizes empirical risk on the training data, researchers employ a \textit{learning algorithm} that iteratively updates the network’s initial weights in a way that successively reduces the loss on the training data. The most common learning algorithms are based on the heuristic of \textit{gradient descent}.\footnote{We will introduce the heuristic of gradient descent learning in section \ref{subsec5.1:FeedbackinNN}.} The final step involves testing the resulting prediction function on the \textit{test set}, a separate set of data points that were not included in the training set. If the prediction function performs well on this test set, it is accepted and usually used for deployment. Otherwise, researchers adjust the hyperparameters and repeat the training process, potentially with a modified hypothesis class.

\label{subsec2.1:NNTraining}

\subsection{Learning Opacity}

The opacity of the NN learning process has already been noted by some authors \parencite[]{lipton2017mythosmodelinterpretability, Boge2022, Sogaard_2023}. For example, \textcite{lipton2017mythosmodelinterpretability} talks about a lack of `algorithmic transparency' in NN and thereby means that researchers do not understand the `heuristic optimization procedures' for NNs (p. 40). \textcite{Boge2022} identifies a type of opacity that `concerns the way in which a DNN automatically alters the instantiated function in response to data' (p. 59).  Similarly, \textcite{Sogaard_2023}  defines `training-opacity' as the inability of expert humans to say, upon inspection, how the parameters of the DNN were induced as a result of its training data (p. 225).
\par 
To delineate our notion of the opacity of the NN learning process from the others, we call it \textit{learning opacity}. We take learning opacity to refer to the lack of understanding of the dynamical phenomena of the NN's \textit{weight dynamics} that appear during the learning process. By weight dynamics we mean the time evolution of the NN's weight values during the training process. To keep our discussion as broad as possible we do not restrict the notion of learning opacity to a specific meaning of `dynamical phenomena' or `understanding'.\footnote{The concept of understanding has been extensively debated in philosophy, and there is no universally accepted definition. For an overview, see \textcite{BaumbergerBeisbartBrun2016, sep-understanding}.}\ Dynamical phenomena in NN weight dynamics may include fine-grained behaviours, such as the evolution of precise weight values, or more coarse-grained behaviours, such as the convergence speed of training, the type of minimum reached, or aforementioned phenomena like saturation or benign overfitting. By not committing to a single definition of understanding we can illustrate the various epistemological challenges that complexity poses for the study of dynamical phenomena of the NN learning process and how these challenges limit different ways of understanding the learning process, thus contributing to learning opacity.\footnote{\textcite{Sogaard_2023} conducts a similar project as he discusses possible reasons for training-opacity. However, his project differs from ours in two important respects: first, he focuses solely on the lack of understanding of how the parameters of a NN are induced by its training data, rather than on phenomena of the learning process in general and second, he does not address how complexity contributes to this lack of understanding.}\ However, before doing so, we first need to clarify what we mean by \textit{complexity}.
\label{subsec2.2:NN_Opacity}

\section{Complexity in the Natural Sciences}\label{sec3:Complexity}

Complex systems in the natural sciences are generally considered to be constituted by many elements that interact in a sophisticated way which results in some
kind of self-organized structure \parencite[cf.][]{WiesnerLadyman, Wiesner_Ladyman_paper, Sandra_Mitchell, Mitchell_Complexity_A_Guided_Tour}.\footnote{The term `complexity', when left unspecified, is used to denote the concept of complexity as it is understood in the natural sciences.} Such structures display some regularity, e.g. symmetry, periodicity or some form of pattern. Roughly speaking, the structure is self-organized in case it arises autonomously out of the interactions among the parts and not from some central control. The self-organized structure in complex systems is a form of emergent behaviour \parencite[cf.][]{Anderson_1972_Moreisdiffernet}.
\par 
Complex systems closely resemble NN systems. NNs are defined as the repeated composition of elements (the neurons) that interact with each other (transmitting information). During training, the strength of the information flow between nodes (the weights) changes in a sophisticated manner, eventually resulting in the optimal weight configuration (the prediction function), a kind of self-organized structure. 
\par 
The resemblance between NNs and complex systems is useful to investigate learning opacity because it directs attention to the dynamic processes and how complex features of the dynamic process limit understanding.\footnote{It is less clear whether complexity concepts from computer science, such as computational complexity \parencite{HartmanisStearns1965} or Kolmogorov complexity \parencite{Kolmogorov1965}, whose systems of study are Turing machines, are suited for the study of learning opacity in NN.} Complexity research in the natural sciences examines not only the complexity of the organized results but also the complexity of the dynamic processes generating them. 
Having the means to examine the complexity of the dynamic processes is crucial for our research question, since we are interested in the complexity of the NN's learning process, i.e.\ the dynamic process that generates the prediction function. Furthermore, there are reflections on how the complexity of the dynamical processes limits our understanding of the system. These provide the conceptual foundation for investigating how complexity contributes to learning opacity in NN learning.
\par 
Yet, what exactly qualifies as `sophisticated' interaction and what kind of self-organized structure is required for a system to satisfy the above characterization of a complex system remains a matter of debate. The first point of contention concerns which qualitative properties are regarded as being necessary and/or sufficient for a system to count as complex \parencite[cf.][]{WiesnerLadyman, HOOKER_Introduction}. Given a specific qualitative property, the second point of contention concerns how it should be quantified \parencite[cf.][]{ Wiesner_Ladyman_paper}. The multiplicity of qualitative and quantitative approaches to complexity reflects both the diversity of systems considered complex and the differing interests of the scientific disciplines studying them. Nevertheless, three qualitative properties appear to be central to most accounts of a dynamic processes' complexity.
\par 
A property that often appears in connection to complexity is nonlinearity and the associated property of sensitive dependence on initial conditions (SI) \parencite[]{BISHOP_Epistemology, HOOKER_Introduction}.\footnote{SI is a defining property of another class of systems, namely chaotic systems. It should be noted that, although many complex systems display SI, chaos and complexity are distinct concepts \parencite[][]{Zuchowski_Disentangeling}.}  
The second property is feedback (F) \parencite[cf.][]{WiesnerLadyman, Sandra_Mitchell}. Lastly, complex systems are often considered to be context-sensitive (CS) \parencite[]{Sandra_Mitchell}. 
\par 
To elucidate how these properties affect researchers’ understanding of dynamical phenomena, we now introduce one of the most common methods used to understand the dynamical behaviour of natural systems: dynamical modelling.

\label{subsec:3.1.}

\subsection{Dynamical Modelling}\label{subsec4.0:Dynmaical Modelling}
Dynamical modelling \parencite[cf.][]{BISHOP_Epistemology, Berger_1998} is a form of mathematical modelling in which natural systems are represented and analysed through comparison with abstract mathematical entities known as \textit{dynamical systems}.
\par 
A dynamical system is a mathematical model that describes how the states of a target system evolve over time. What counts as the relevant states of the target system depends on the researcher’s explanatory goals. The most accurate and fine-grained approach characterises the evolution of the target system in terms of the states of its components, the system’s \textit{micro-states}. The \textit{state space} of a dynamical system comprises all possible states the system could, in principle, occupy.
\par 
The time-evolution of the target system is modelled via a \textit{time evolution function} which is typically derived from \textit{dynamical equations} that specify how the system’s state changes over time. These equations take the form of difference equations in the discrete-time case, or differential equations in the continuous-time case. Together with the \textit{initial conditions}, that describe the system’s state at the start of its evolution, and the \textit{boundary conditions}, that describe the constraints on the system, the dynamical equations uniquely determine the time-evolution function. A \textit{trajectory} represents a particular path that the system follows through the state space as it evolves over time. 
\par 
We here propose to view the NN's weight dynamics during training as a dynamical system. The micro-states correspond to the weight values, the time evolution is governed by the learning algorithm, and the external conditions, the context, are given by the training data (see Table \ref{tab:ComparisonDynamical})\parencite[cf.][section 4.1]{Levin_Dis}. The dynamical systems perspective on NN learning provides a fruitful framework for analysing how SI, F, and CS contribute to learning opacity, which will be addressed in the remainder of the paper. For each property, we first discuss its characterization in the natural sciences and assess whether it is also present in NN learning. We then examine its impact on the understanding of complex systems via dynamical modelling, and whether it contributes to learning opacity in a comparable manner.
\begin{table}[h!]
    \centering
    \renewcommand{\arraystretch}{1.8}
    \begin{tabular}{P{4cm} P{5cm} P{5cm}}
        \toprule 
        & \textbf{Complex systems} & \textbf{NN learning} \\
        \midrule\midrule
        
        \textit{Micro-states} 
        & States of constituents 
        & Weight values \\
        
        \textit{State space} 
        & All possible states of constituents 
        & All possible weight values \\
        
        \textit{Initial states} 
        & Initial states of constituents  
        & Initial weight values \\
        
        \textit{Context} 
        & Environment 
        & Training data \\
        
        \textit{Time evolution} 
        & Differential or difference equations governing micro-state evolution 
        & Weight update algorithm \\
        
        \bottomrule
    \end{tabular}
    \caption{Analogy between complex systems and neural network learning understood as dynamical systems.}
    \label{tab:ComparisonDynamical}
\end{table}

\section{Sensitive Dependence on Initial Conditions}\label{sec:SI}

\textit{Sensitive dependence on initial conditions} (SI) means that small changes in initial states might be amplified over time and lead to significantly different system behaviour.
\par 
In dynamical systems the phenomenon of SI is closely associated to nonlinearity \parencite[]{BISHOP_Epistemology}. A function $f$ is \textit{linear} if it satisfies both of the following conditions:\footnote{The first condition is called \textit{superposition} and the second \textit{homogeneity}.}
\begin{align*}
    & f(x+y) = f(x) + f(y)\\
    & f(cx) = c f(x), \text{where $c$ is a constant.} 
\end{align*}
A \textit{nonlinear} function violates at least one of these conditions. In the context of dynamical systems, a nonlinear system is a system whose time-evolution is nonlinear. Unlike linear functions, a change in the input of a nonlinear function does not necessarily produce a proportional change in the output. So, in a nonlinear dynamical system even small changes in the input may be amplified over time, leading to drastically different outcomes.
\par 
%More formally, SI of a map $f$ means that that there is a $\delta > 0$ such that for any initial point $x$ and any ball of of radius $\epsilon$ around $x$ there exists $y$ in this ball and time $t$ such
%that for the systems state at time $t$, $f_t$, the property $|f_t(x) - f_t(y)| > \delta$ holds. Thus, in systems that have SI there can be exponentially fast separation of nearby trajectories for
%infinitesimally changed initial conditions. 
\par 
SI is commonly understood as the exponential growth of errors signified by a positive global Lyapunov exponent. A positive global Lyapunov exponent measures the average exponential rate of divergence of infinitesimal perturbations of initial conditions in a dynamical system over infinite times. However, the global Lyapunov exponent is criticised as an adequate measure of SI \parencite{BISHOP_Epistemology}. A positive  Lyapunov exponent only measures the \textit{average} effect of \textit{infinitesimal} perturbations for an \textit{infinite} time period. Finite deviations are not guaranteed to exhibit average growth rates for a finite time at the rate predicted by the Lyapunov exponent.
Hence, a positive Lyapunov exponent might not adequately capture a possible rapid growth of \textit{finite} deviations in \textit{finite} time. We therefore don't consider a positive Lyapunov exponent as a necessary condition for SI. Instead, we adopt a qualitative understanding, according to which SI means that small changes in initial states might eventually lead to significantly different system behaviour.
\par 
An example of a system exhibiting SI is a ball placed on the peak of a gable roof. Whether the ball rolls down the right hand side of the roof or the left hand side can be determined by very small displacements of its initial position. The first thorough scientific examination of a system that is sensitive to small perturbations in its initial conditions was done by Edward N. Lorenz discussing atmospheric convection phenomena \parencite{LorenzDeterministicNonperiodicFlow}. A famous one-dimensional map in the interval $[0,1]$ that is used to study SI is the logistic map $x_{n+1}=rx_{n}(1-x_{n})$ popularized by the biologist Robert May \parencite{May}.
Population biologists examine the logistic map to model the effect of the population size $x_n$ in generation $n$ on the next generation's population size $x_{n+1}$, where $r$ is interpreted as the population's growth rate. Small differences in the initial conditions of the logistic map already have a huge effect after a small number of iterations. For $r=4$ and the initial value $x_0 = 0.51$ the map outputs $x_{10} = 0.53049$ after 10 iterations. However, starting with initial value $x_0= 0.52$ yields $x_{10} = 0.99577$. 

\label{subsec:3.4.}

\subsection{Weight Initialization Sensitivity in NN Learning}\label{subsec5.2SIinNN}

A NN's learning process displays SI in the sense that its behaviour strongly depends on the strategy used to initialize the weights. 
\par 
In the natural sciences, an initial condition is the state of a system at the beginning of the time period under investigation. Models used to study complex systems frequently exhibit SI, since complex natural systems are often best represented by models with nonlinear dynamics. NNs also exhibit nonlinearities, namely their nonlinear activation functions. In NNs, however, it is not the precise weight values at initialization that matter most, but rather the \textit{weight initialization strategy}. The general approach to weight initialization is to select a probability distribution and draw the initial weight values as samples from it. Specific strategies differ in their choice of distribution and in how the parameters of that distribution are set. Several results demonstrate that nonlinear activation functions interact with the initialization strategy in ways that affect convergence speed and generalization performance \parencite[cf.][]{He_initialization, understanding_Glorot, Initial_Conditions_Generalization_Performance}.
\par 
\textcite{understanding_Glorot} show that for certain activation functions the initialization strategy strongly influences the number of neurons that are saturated at initialization and during training. A neuron is said to be \textit{saturated}, when its input lies in a region of the activation function where the activation function's output changes very little with respect to changes of the neuron's value. Saturated neurons impact learning behaviour because the magnitude of a weight's update step is proportional to the sensitivity of the activation function to changes in the neurons' value that contain that weight. Thus, weights involved in saturated neurons may update only very slowly. If many neurons are saturated, the training process may converge extremely slowly or fail to converge at all. 
\textcite{He_initialization} provide direct empirical evidence that weight initialization strategies can decisively influence convergence speed. They even show that in one classification task a NN initialized with one weight initialization strategy fails to converge, while the same network initialized with another weight initialization strategy converges. As for generalization performance, \textcite{understanding_Glorot} demonstrate that there is at least one activation function, the hyperbolic tangent,\footnote{The hyperbolic tangent activation function, tanh($x$), has the form $\frac{e^{x}-e^{-x}}{e^{x}+e^{-x}}$.} for which the initialization strategy influences the test error. For this activation function the difference in test error between different initialization strategy can rise up to 12\%. 
\begin{comment}
\par 
\textcite{He_initialization} show difference in convergence speed between the normal Xavier initialization strategy and their proposed He initialization strategy for the rectifier activation function. Most prominently, the show that in one classification task the NN initialized with Xavier initialization doesn't converge at all, in contrast to the NN initialized with He initialization.
\footnote{See figure 11 in \textcite{understanding_Glorot} for the difference in convergence speed and performance of the two weight initialization strategies for the tanh activation function. Figure 11 shows the results for the Shapeset training set. \textcolor{red}{The experiments were also conducted for other image recognition sets.}} \textcolor{red}{Also talk about non-linearity?}
    \footnote{\textcolor{red}{Two phenomena that cause reduction in convergence speed are saturation of activation functions and vanishing gradients \parencite[]{Sepp_Hochreiter_Vanishing_Gradients, Bengio_Expldoing_Gradients}. Whereas vanishing gradients are more problematic for recurrent neural networks as for feed-forward NN \parencite{sussillo2014vanishinggradient}, saturation is a hard problem for NN training. Or should I leave this comment and explain in th etext what saturation of activation functions is and how it is connected to slow learning? I have no goog paper that investigates vanishing gradients in FFNN.}} 
\end{comment}
\par 
Arguably, whether these findings show that the training process of a NN is sensitive to initial conditions, meaning that even a small change in the initialization strategy can lead to a significant difference in weight dynamics, depends on what counts as a small change in the initialization strategy and what counts as a significant difference in weight dynamics.
\par 
The weight initialization strategies compared in \textcite{understanding_Glorot} are both uniform distributions that only differ in their parameters. Likewise, the initialization strategies examined in \textcite{He_initialization} are Gaussian distributions that also differ in just one parameter. We think that a difference only in the parameters of the probability distribution used to sample the initial weights can be reasonably considered to be a small change in the initialization strategy. Regarding the effects on weight dynamics, a difference between converging behaviour and non converging behaviour constitutes a significant effect. A difference of about 12\% in test error indicates that the respective weight dynamics ended up in quite different minima.\footnote{For comparison: the ImageNet Large Scale Visual Recognition Challenge \parencite[]{2015imagenet}, the standard benchmark for image recognition, saw a total improvement of 13.7\%, from AlexNet's 16\% error rate in 2012 to SENet's 2.3\% in 2017, when the contest ended. A 12\% drop thus represents nearly the entire progress made by the field over that 5 year period.}

\subsection{Sensitive Dependence on Initial Conditions and the Limits of Representability}\label{subsec4.2:SIandNon-Predictabiliyt}

In natural complex systems, SI makes it extremely difficult to construct an adequate representation of a system’s dynamics, often forcing researchers to rely on empirical investigation and statistical inference. This, in turn, challenges our understanding of the system by introducing the limitations inherent in statistical analysis.
\par 
In the dynamical modelling approach, the model used to represent the dynamics of the target system is the dynamical system consisting of the time-evolution function and the state space. If the time-evolution function of the dynamical system exhibits SI, even small variations in the initial states can eventually lead to drastically different outcomes. However, it is generally impossible to measure the states of a natural system with arbitrary precision \parencite[cf.][]{Crutchfield1994}. Consequently, the rapid amplification of errors in the real initial conditions of the target system imposes severe limitations on our ability to use the dynamical system to predict the target system's evolution in time.
\par 
The challenges of using the dynamical system to predict the target system's evolution makes it difficult to confirm the dynamical model, discrete or continuous, such that the researchers are confident that the model adequately captures the target system’s dynamics. This is because SI poses serious challenges to the common strategy of piecemeal confirmation.\footnote{Other methods for model confirmation are available, but encounter problems similar to those faced by the piecemeal improvement strategy when applied to models exhibiting SI  \parencite[cf.][]{BISHOP_Epistemology, Koperski_model}.} There are two strategies for confirming a model in a piecemeal fashion. One approach is to increase the accuracy of the initial and boundary conditions while keeping the model itself fixed \parencite{Laymon1989}. Alternatively, researchers may refine the model incrementally, e.g.\ by de-idealizing it, while keeping the initial and boundary conditions constant \parencite{Wimsatt_successive_model}.
\par 
Assuming that the model correctly represents the target system, a successive increase in the accuracy of the initial conditions should lead to a monotonic convergence of the model’s predictions toward the actual behaviour of the target system. Similarly, in the second strategy, if the initial model captures the target system’s mechanisms to some extent, successive refinements should produce increasingly accurate predictions. While monotonic convergence is taken as evidence for the initial model’s adequacy in representing the target system’s mechanisms, a failure to converge suggests that the initial model may not adequately capture those mechanisms.
\par
The monotonic convergence approach to confirmation, however, faces great difficulties if the model deploys SI \parencite{Koperski_model, BISHOP_Epistemology}. In the first case, an increase of the accuracy of the initial conditions of the model means that the initial conditions are slightly changed. This might lead to an extreme change of the predicted trajectory due to SI. It is not guaranteed that the resulting prediction is closer to the target system behaviour, even when the new, improved initial conditions more closely approximate the actual initial conditions of the target system. Moreover, the amplification of measurement error and researchers' inability for accurate measurements blocks the second convergence strategy, fixing the data and improving the model. The problem is that \textit{even if} researchers had the correct (non-linear) time-evolution function, they would never be able to know whether it is the correct time-evolution function if it deploys SI, since they would never be able to have infinitely precise initial conditions. However, if the dynamical model with the correct time-evolution function failed to predict the target system’s true behaviour, then neither will any of its refined versions, and it may be the case that researchers never attain sufficient confidence in the adequacy of any of the models.
\par 
When researchers are unable to study a target system's dynamics using an adequate dynamical model, they are compelled to investigate it experimentally, that is, through direct observation. Reliance on direct observation marks a shift toward a mode of reasoning grounded in statistical inference. Researchers examine the natural system's dynamics by conducting repeated experiments, reporting the results and drawing general conclusion from the collected data in an inductive fashion. Arguably, the frequentist approach remains the most widely used method for inductive inference in scientific practice. Yet, it is very challenging to actually implement in practice the theoretical assumptions required for a valid inference from frequentist methods. Moreover, even if the researchers correctly apply frequentist methods they sometimes misinterpret the results, which leads to mistaken conclusions from the empirical data. All this contributes to the fragile nature of hypotheses about the system’s micro-states dynamics when they are based on empirical data and inductive inferences \parencite[cf.][]{Sprenger_Frequentism_vs_Bayesianism, Schneider_Problems_of_Statistics}.

%Difficulties in constructing an adequate model of a target system limit our understanding, since possessing such a model plays a central role in gaining insight into the system’s behaviour. The importance of models for understanding is evident in their ubiquitous use across all branches of science. Recent philosophical accounts of scientific understanding emphasize this crucial role of models in facilitating scientific understanding \parencite[cf.][]{understanding_scientific_understanding, Morrison_1999, Knuuttila2009}. De Regt (\citeyear{understanding_scientific_understanding}) maintains that a phenomenon is scientifically understood by integrating it in a theory through the construction of a specific model that represents the phenomenon in a way that allows theoretical principles to be applied (p. 32). Yet, a model can also facilitate understanding independently of a theory. \textcite{Morrison_1999} views models as autonomous agents in knowledge production that mediate between theories and the world. Nevertheless, like de Regt, she emphasizes the importance of models for understanding the target system, asserting that ``[\dots] when we want to find out just how a physical system actually functions we need to resort to models to tell us something about how specific mechanisms might be organised to produce certain behaviour.'' (p. 43) The challenge to identify an adequate dynamical model hinders a model-based understanding of the target system’s dynamical phenomena.

\subsection{Weight Initialization Sensitivity and Opacity}\label{subsec:6.2ContrWeightInit}

Sensitivity to the weight initialization strategy does not contribute to learning opacity by  complicating the construction of an adequate representation of the target system and thus does not force a shift to empirical studies. Rather, it limits the researcher’s pragmatic understanding of the learning process.
\par 
Unlike in complexity science, researchers in ML have full access to the target system. The learning algorithm itself provides a fully accurate representation of the weight dynamics during training and as long as researchers studying the NN learning process have access to the learning algorithm, the hyperparameter settings, the training data and the source code they have access to a complete representation of the learning dynamics.\footnote{This scenario can be considered to be what Bordt and co-authors call a `fully transparent scenario' \parencite[]{Bordt_Facct}.}
\par 
Nevertheless, sensitivity to weight initialization affects the pragmatic understanding researchers have of the learning process. A scientist has pragmatic understanding of a target system insofar as she is able to recognise qualitatively characteristic consequences of the assumptions of the target system’s representation without exact calculations and can successfully intervene on and control the target phenomenon \parencite[]{Lenhard2009, Lenhard_Calculated_Surprise}.\footnote{The idea that understanding has to do with anticipating qualitative characteristics is inspired by the notion of `intelligibility' \parencite[cf.][]{understanding_scientific_understanding, deRegtDieks2005}.} Arguably, because the learning algorithm itself can be considered as a fully accurate representation of the target system in the context of NN learning, one possesses pragmatic understanding of a NN's learning behaviours if one can design the NN learning process in such a way that the NN exhibits desired qualitatively characteristic learning behaviours. However, it is typically very difficult to anticipate even qualitatively the consequences of a given weight initialization strategy for the resulting learning dynamics prior to training. Consequently, there exist very few principled guidelines for selecting weight initialization strategies that yield a desired learning behaviour and determining the `correct' weight initialization is often a matter of guessing \parencite[][p. 1]{weight_initialization_guess}.\footnote{This observation holds for hyperparameter settings in general, as NN modellers often need to first guess the best setting and then fine-tune it through trial and error.}

\section{Feedback}\label{sec:Feedback}

\textit{Feedback} (F) means that there are mutual dependencies between different parts of the system. Yet, there are different ways of conceiving the parts involved in such dependencies: feedback may occur from past interactions to current element interactions \parencite[]{WiesnerLadyman}, between elements and their interactions \parencite{Thurner_Complex_Systems, Holovatch_2017}, or between micro-level properties and system-wide properties \parencite{Estrada_Whatisacomplexsy, Sandra_Mitchell}.
\par 
These mutual dependencies generate feedback loops, where the parts involved continuously influence one another. Importantly, feedback tends to blur the causal order. There is no simple, one-way causal chain from one part of the loop to another. This also means that the parts that constrain the behaviour of others are themselves influenced by the very behaviours they constrain. In particular, feedback challenges the traditional bottom-up conception of causation, in which causal influence flows only from lower levels to higher levels. In complex systems, micro-level behaviour generates emergent system-wide properties, which in turn feed back to constrain the micro-level dynamics \parencite[pp. 34]{Sandra_Mitchell}. 
\par 
A classic example of a system with feedback is the predator-prey interaction described by the Lotka–Volterra model \parencite[]{volterra1926variations, lotka1925elements}. This model formalizes the mutual dependency between predator and prey populations and shows that these two variables fluctuate in a periodic manner. The Lotka-Volterra model illustrates well the interaction between micro-level dynamics and constraining system-wide properties characteristic of feedback systems. The micro-level dynamics of successful hunting attempts of an individual predator or the successful reproduction of an individual prey influence both the sizes of the predator and prey populations, two system-wide properties. Yet the predator and prey population sizes, in turn, constrain the micro-level interactions of successful hunting and successful reproduction.
\par 
\vspace{5mm}
%An example of a system exhibiting feedback is a model of voting behavior \parencite{Holovatch_2017}. This is an example of a system with feedback between type of interaction and node value. (\textcolor{red}{Important example, because the type of feedback I describe in NN is between nodes and links.}) Voting behaviour - no social contact no change. High influence of family little change to change to a different opinion, high effort. Voting behaviour: polls - the more people vote for one party the more we think this party is good. However, they interaction between two people has influence on the overall polls and so influences system wide property.

\label{subsec:3.3.}

\subsection{Feedback in NN Learning}In supervised NN training there is a feedback loop between the micro-level dynamics of individual weight updates and the NN loss gradient, which is a system-wide property.

%\footnote{On a broader picture there is also feedback between the NN system and the data used to train it \parencite[cf.][]{Sebastian_Zezulka, Human_AI_Coevolution}. However, we focus on the feedback between weights and nodes, because this feedback is inherent to gradient descent updating. Is there feedback between the hyperparameter of learning rate and the weights? Or do we set the learning rate independent of the weights?}
\par 
In supervised learning, we train a NN using a training algorithm that finds, based on labelled data, a NN configuration that can serve as a good prediction function on future data points. Currently, the most widely used training algorithms for supervised learning are based on gradient optimization.\footnote{Here, we illustrate gradient-based optimization in the case where only a single data point is used to update the network at each step. For a general introduction to gradient-based optimization, see \textcite[ch. 8] {Goodfellow-et-al-2016}.}
\par 
A derivative of a function $f(x)$ is the slope of $f(x)$ at point $x$. If a function $f(x,y)$ depends on two variables $x$ and $y$ we call the derivative with respect to a single variable the \textit{partial derivative}. The concept of a derivative generalizes to a gradient in the multi-dimensional setting. A \textit{gradient} is a vector containing all the partial derivatives of the respective function. The negative gradient of a function $f$ at point $x$ points to the steepest descent of $f$ at $x$. The basic idea of gradient-based optimization is to use the gradient to iteratively update the weights and biases of the NN, thereby finding a configuration that yields a low loss on the training data.
\par 
The loss function $\mathcal{L}$ maps the current NN, $f_{w}$, and a data point, $(x^{(i)}, y^{(i)})$, to the corresponding loss value $\mathcal{L} \big(f_{w},x^{(i)}, y^{(i)} \big)$ and thus depends on the weight variables $w_i$ in $w$. In the learning process, the researcher's goal is to find a weight configuration such that the corresponding NN minimizes the loss on the training data. For any given data point, the negative gradient of the loss function with respect to the network weights points in the direction in which the weights must be updated in order to minimize the loss most effectively. By successively updating the weights of the network in this direction, we gradually approach a configuration with lower loss. 
\par 
Schematically, one update step in gradient-based optimization for an individual weight value $w^{(n)}_i$ at step $n$ is given by:
\begin{equation} \label{eq:gradient_learning}
    w_{i}^{(n+1)} = w_{i}^{(n)} - \epsilon \Big(g_{w}\mathcal{L}\big(w^{(n)},x^{(i)}, y^{(i)}\big)\Big)_i.
\end{equation}
We subtract from the weight value $w_{i}^{(n)}$ the value of the $i^{th}$ entry of the loss function's gradient $g_{w}\mathcal{L}\big(\cdot, \cdot, \cdot\big)$,\footnote{The gradient $g_{w}\mathcal{L}(\cdot, \cdot, \cdot)$ is the vector in which the $i^{th}$ entry, $\big(g_{w}\mathcal{L}(\cdot, \cdot, \cdot)\big)_i$, is the derivative of $\mathcal{L}(\cdot, \cdot, \cdot)$ with respect to $w_{i}$.} evaluated on the current NN weight configuration $w^{(n)}$ and the current data point $(x^{(i)},y^{(i)})$. This scalar value is scaled by the learning rate $\epsilon$. 
\par 
Equation \ref{eq:gradient_learning} shows that the size and direction of an updating step of an individual weight is constrained by the $i^{th}$ entry of the evaluated loss gradient $g_{w}\mathcal{L}\big(w^{(n)},x^{(i)}, y^{(i)})$. This scalar in turn depends on the general form of the loss gradient $g_{w}\mathcal{L}\big(\cdot, \cdot, \cdot\big)$, which is calculated for the specific NN architecture. Thus, the updating step of an individual weight depends on the distribution of weights across the whole NN. Moreover, to update an individual weight the gradient is evaluated for the current weight value configuration, that means that \textit{all} current weight values potentially affect an individual weight's updating step. The dependence of the gradient value on the architecture and the whole weight value configuration makes it a system-wide property. 
\par 
Conversely, because the gradient is evaluated with respect to the overall weight configuration, a change in an individual weight can affect the value of the gradient. This implies that micro-level dynamics, individual weight updates, can influence the macro-level dynamics of the gradient.
\par 
In sum, the micro-level dynamics of individual weight updates influence the overall NN weight configuration and thus the system-wide property of the loss gradient value, which in turn constrains the further weight updates. 

\label{subsec5.1:FeedbackinNN}

\subsection{Feedback and Computational Modelling}\label{subsec4.1:FedandNonRepre}

In natural complex systems, F prevents the system's micro-state time evolution from being representable in a tractable and theoretically useful way. As a consequence, F limits researchers' understanding of the target system’s behaviour, as they instead study its dynamics through an incomprehensible model.
\par 
As mentioned above, the most accurate and fine-grained way to represent a system’s state evolution is by modelling the dynamics of its constituents. Classically, one would represent the system’s dynamics by deriving an \textit{analytic} time-evolution function for these micro-states \parencite[cf.][]{PARISI}. Deriving such a function requires solving the corresponding differential equation under fixed constraints on the system, expressed as constant boundary conditions.
\par 
In complex systems, however, the constraints on the systems dynamics are not fixed.
This is because feedback occurs not only between the system’s components but also between the components and the very conditions that constrain their behaviour. Recall the example of the Lotka-Volterra model. The overall prey population constrains the number of successful hunts of an individual predator. Yet, the number of the predator's successful hunts influences the overall prey population, which in turn influences the number of successful hunts of the individual predator. The feedback between the system's components and their constraints has thus the consequence that there are no constant boundary conditions and so it is often impossible to derive an analytic time-evolution function that represents the dynamics of the complex system's micro-states in a theoretically useful way \parencite{Holovatch_2017, Thurner_Complex_Systems, HOOKER_Philosoph_of_Complex_Systems}.\footnote{Note that this does not imply that it is impossible to describe analytically the dynamics of more \textit{coarse-grained} states of the system. In the the Lotka-Volterra model for example researchers are able to describe analytically the time-evolution of the predator and prey \textit{populations}.} 
\par 
Nevertheless, researchers studying target systems exhibiting feedback might have access to dynamical equations that are intended to represent the evolution of the system’s microstates. Yet, as argued, these equations often lack analytic solutions. In such cases, the standard approach for investigating the system’s behaviour is to rely on computer simulations. Researchers may either approximate the solution of analytically intractable differential equations computationally, or directly simulate the target system using discrete difference equations from the outset. The resulting simulated behaviour can then be treated as a model of the target system for the purpose of studying its dynamics \parencite[cf.][]{Weisberg_Computer_Simulation_as_model}.
\par 
Computer simulations are a source of uncertainty, since computers are finite, discrete machines and therefore subject to rounding errors. Moreover, there are multiple ways to implement a given simulation, and the resulting behaviour may depend on the specific implementation details \parencite[p. 37]{HOOKER_Introduction}. At a more philosophical level, it remains unclear how computer simulations contribute to our understanding of a target system’s behaviour. One of the most widely discussed issues concerning the epistemic status of computer simulations is their epistemic opacity \parencite[cf.][]{Humphreys2009ThePN, Duran_Computer_Simulations, Humphreys2004}. Epistemic opacity arises from the fact that human agents, as finite reasoners, cannot have knowledge of all the epistemically relevant steps involved in producing a simulated state description of the target system \parencite[]{Humphreys2009ThePN}. What counts as an `epistemically relevant' step in the use of a computer simulation, however, is often vaguely defined \parencite[cf.][]{Beisbart2021-Opacity_CS}. On a broad conception, epistemically relevant steps may span the entire modelling process, including modelling assumptions such as parameter choices \parencite[cf.][]{WhyTrustSimulation}. On a narrower conception, they may be restricted to the computational steps internal to the simulation itself \parencite[cf.][]{Humphreys2009ThePN}. Whatever one takes to be the epistemically relevant steps, the epistemic opacity of computer simulations challenges researcher's ability to explain how exactly simulated phenomena, such as emergent macro-level feature, are generated by the computer simulation. Understanding the behaviour of the real-world target system becomes thus even more difficult, since the model that is used as the target system's representation is not comprehensible \parencite[cf.][]{Humphreys2009ThePN}.

\subsection{Feedback and Opacity}

We now argue that the challenge posed to ML research by the absence of a tractable and theoretically useful representation does not lie in the use of incomprehensible models, but rather in the challenges of statistical inference.
\par 
Although ML research is considered to lie at the interface of a formal and empirical science \parencite[cf.][]{Rethink_empirical_ML}, NN researchers struggle to find a tractable and theoretical useful representation of the fine-grained weight dynamics of realistic NN during learning \parencite[cf.][p. 1]{ Cohen_understanding_central_flow_ICML}. Given the affect of feedback in natural systems on the possibility of deriving a theoretical representation, feedback loops in gradient-based optimization might be a source for this. The difficulty in theoretically analysing the learning behaviour of realistic NNs significantly complicates the first type of inquiry and has led to an increased reliance on empirical approaches in ML.
\par 
As pointed out in section \ref{subsec:6.2ContrWeightInit}, ML researchers have full access to the target system. An ML experiment consists in executing the learning process on a computer and directly observing how it unfolds. Since the learning process is implemented on computers, uncertainties due to interaction effects between the hardware and the program, like rounding errors or the form of implementation, do play a role \parencite[]{Creel_2020}. Yet, due to the full access to the true dynamical description of the target system, empirical ML researchers are not required to represent or approximate the behaviour via a computer simulation.\footnote{
We are not claiming that the features that render computer simulations epistemically opaque cannot also arise in the execution of NN learning processes on a computer. However, we think that these problems are not generated by the complex properties that are the focus of the present paper.} 
\par 
The reliance on direct observation in empirical ML has imported an inductive perspective such that researchers conduct (repeated) experiments and try to generalize their insight via statistical reasoning \parencite[]{Rethink_empirical_ML}. Thus, the challenges that arise due to the lack of a tractable and theoretical
useful representation of the fine-grained weight dynamics of realistic NN are the epistemological challenges of empirical testing and statistical inference mentioned in section \ref{subsec4.2:SIandNon-Predictabiliyt}. \textcite{Rethink_empirical_ML} identify four methodological challenges that the inductive perspective in ML research encounters. First, the experiments in empirical ML are often biased in favour of the results the researchers want to demonstrate. Second, there is an unilateral focus on performance experiments. Third, the important abstract concepts in ML are often not well defined and it is not clear how they should be operationalized in experiments. In addition to the three problems related to experimental design, empirical ML encounters the aforementioned problem that the theoretical assumptions underlying statistical tests are rarely satisfiable in practice, and the reliability of statistical test results is frequently overstated. The authors argue that if researchers adopt the inductive perspective uncritically, these problems tend to go unnoticed, and the results end up being non-reproducible, as has already happened in practice \parencite[sec. 1]{Rethink_empirical_ML}. The non-replicability of certain results regarding the NN behaviour drawn from empirical studies suggests that the above epistemological challenges are hindering the understanding of the NN learning process, thus contributing to learning opacity.

\label{subsec:6.1ContrFeed}

\section{Context Sensitivity}\label{sec:CS}

\textit{Context-sensitivity} (CS) refers to the dependence of the system’s behavior on conditions in its environment \parencite[cf.][]{Sandra_Mitchell}. In other words, a context-sensitive system is influenced not only by its internal parts and structure but also by external conditions. 
\par 
In physics, context-sensitivity is understood as a system being open. An open system is a system with a net influx of energy or matter to the system. This influx usually originates from outside the system, which makes open systems driven by something external. Open systems are generally not in thermodynamic equilibrium, though, some open systems with a compensating outflux can reach a dynamic equilibrium in which certain system properties remain approximately constant over time. How the physical notions of open system and thermodynamic equilibrium are generally applicable to non physical systems, like financial markets, remains an open question \parencite[]{WiesnerLadyman}.
\par
An example of a natural system that is context sensitive is the foraging behaviour of ants \parencite[]{Ants_Foreaging}. Researchers have found that, in some ant species, foraging behaviour is influenced by environmental temperature. Specifically, the ants shift from a unimodal foraging pattern, that is, continuous foraging throughout the day, to a bimodal pattern, with two distinct foraging periods separated by a break, in order to avoid the high midday temperatures. Furthermore, when a system is subjected to multiple external influences, the order in which these influences arrive matters. For example, a plant's successful growth depends on more than just eventually receiving water and sunlight. Receiving excessive sunlight without sufficient water will dry the plant out, whereas too much water without adequate sunlight will drown it.

\label{subsec:3.5.}

\subsection{Context Sensitivity in NN Learning}The learning dynamics of a NN is context sensitive, because its dynamics are sensitive to the training data distribution and order. 
\par 
For NNs, the training data are external conditions and thus its context: the data is generated in the world and processed by the researchers and not part of the network’s formal definition. Importantly, the dynamics of learning are shaped both by the training distribution itself and by the order in which training examples are presented.
\par 
That the training data distribution influences learning dynamics is a desired property. Otherwise, learning different distributions would be impossible. A clear illustration of this effect is provided by \textcite{zhang2017understanding}, who trained a NN for image classification. They showed that altering the correlation between images and labels, i.e.\ modifying the training distribution, directly impacts convergence speed and generalization performance.
\par 
The sensitivity of learning dynamics to the order of training data presentation is apparent in \textit{curriculum learning} \parencite[cf.][]{ELMAN199371, Soviany2022_Curriculum, Wang_Curriculum_Learning}. In curriculum learning, researchers reweight the training distribution over time, making certain examples more likely to appear at specific stages of training. This imposes an order on the training data. The process typically relies on two components: a difficulty measure that assigns a difficulty score to each training example, and a training scheduler that controls how examples of varying difficulty are presented over the course of training.
\par 
\textcite{Bengio_Curriculum} provide an early demonstration of curriculum learning in an image classification task with a NN trained via online stochastic gradient descent.\footnote{In online stochastic gradient descent, gradient optimization is performed on one randomly chosen training example at a time \parencite[ch. 8] {Goodfellow-et-al-2016}.} The task was to classify shapes into three categories: rectangles, ellipses, and triangles. The authors defined difficulty using shape variability. The dataset \texttt{GeomShapes} contained rectangles, ellipses, and triangles with high variability, while \texttt{BasicShapes} contained images of squares, circles, and equilateral triangles with less variability. Thus, the task to classify images in \texttt{BasicShapes} is a supposedly simpler task than clssifying the images in \texttt{GeomShapes}. Accordingly, the curriculum strategy was to first train on \texttt{BasicShapes} before switching to \texttt{GeomShapes}. Bengio and co-authors found that generalization performance was sensitive both to the time split between the two datasets and to the use of curriculum learning at all.

\par 
The sensitivity of the learning dynamics to the order of data is not specific for online stochastic gradient descent nor limited to generalization performance. \textcite{power_of_curriculum_learning} investigate curriculum learning in the context of mini-batch stochastic gradient descent. This is a technique, where stochastic gradient descent updates are performed on small, randomly chosen subsets of training data, the `mini-batches' \parencite[ch. 8] {Goodfellow-et-al-2016}. To impose an order on the training data, the idea is to sample the mini-batches non-uniformly from the training set and regulate the type of batches that are presented to the NN in the mini-batch stochastic gradient descent. 
\par 
The authors show that different curriculum strategies, defined by different difficulty measure and different training scheduler, affect not only generalization performance but also convergence speed. In their experiments with convolutional NN for classification, the convergence behaviour varied substantially depending on the curriculum strategies, that is on the imposed training data order.
\par 
\begin{comment}
The researchers first order the examples in the whole training set by some difficulty measure.
%\footnote{The authors use the confidence score of a classifier to rank the examples.} 
They then use this order to create subsets of the whole training set that differ in the difficulty of the examples they contain. The process of successively training the NN with examples with a higher difficulty is established by first sampling the mini-batches from subsets with simpler examples and continuously increasing the number of difficult examples in the subset the mini-batches are sampled from. Thus the probability of encountering more difficult training data during training is successively increased.
\end{comment}

\begin{comment}
\footnote{\textcite{power_of_curriculum_learning} call the function used to assign a difficulty score to the training data ``scoring function'' and the function that determines the schedule is the ``pacing function''.}
\end{comment}

\label{subsec5.3:CSinNN}

\subsection{Context Sensitivity and Reduced Unification}\label{subsec4.3:SCandUnification}In natural complex systems, context-Sensitivity (CS) poses a challenge for understanding because it reduces the degree of unification in explanations.
\par 
\begin{comment}
    \footnote{\textcolor{red}{Mitchell Paper Exporting Causal Knowledge argues for this claim in the specific case where we want to generalize causal knowledge add it!!!.}}
\end{comment} 
Unifying explanations of a phenomenon contribute to understanding by allowing us to view the phenomenon as an instance of a greater scheme rather than as an isolated event \parencite[cf.][]{understanding_scientific_understanding}. There are two basic strategies by which unification can be achieved. Michael \textcite{Friedman_unification} maintains that explanations consist of derivations of the phenomenon presented in the form of arguments. He argues that the more comprehensive the generalization used in the argument is, i.e.\ the more situation it subsumes, the more unifying the explanation becomes.
Like Friedman, \textcite{Kitcher1981, Kitcher1989} also considers explanations to be arguments. Yet, he argues that the unifying power of an explanation lies not in the use of a comprehensive generalization, but in the application of an argument pattern that is as general as possible. Two phenomena are unified in case their explanations instantiate similar argument patterns.
\par 
In both accounts, CS limits our capacity to provide unifying explanations. In general, the truth of a generalization is contingent on background conditions, and the degree of this contingency depends on the stability of the conditions under which the generalization holds \parencite{Mitchell_2000, woodward2000explanation, woodward2010causation}.\footnote{\textcite{woodward2000explanation} calls a generalization `invariant' if it would remain stable as various other conditions change. The range of changes over which the generalization is invariant is the `domain of invariance' (p. 205).} Whether a generalization can be applied to explain a target system's behaviour depends on whether the necessary conditions under which the generalization holds true are satisfied in that specific case. In scientific disciplines that investigate systems that are influenced by many external conditions to a high extent, such as biology, the stability of the relevant background conditions is presumably lower than in disciplines that study less context-sensitive systems, such as physics \parencite[cf.][]{ Mitchell2002_lessonsfrombiology, woodward2010causation}. Consequently, in disciplines that study highly context-sensitive target systems, the generalizations employed in explanations are less general, as their conditions are less often satisfied in concrete cases. 
\par 
Moreover, it is plausible to assume that more argument patterns are required when addressing context-sensitive phenomena. Argument patterns are abstracted from the particular arguments we use to derive, that is, to explain, the phenomenon in question \parencite[p. 520]{Kitcher1981}. However, different external conditions might give rise to different phenomena, which might require distinct arguments for their explanations. The greater the number of arguments required to account for the observed phenomena, the more likely it is that these arguments instantiate different argument patterns. Hence, the unifying power of a concrete explanation that instantiates a single general argument pattern is diminished.

\subsection{Training Data Sensitivity and Opacity}As with sensitivity to weight initialization, sensitivity to training data affects the researcher's pragmatic understanding of the learning process. The extent to which sensitivity to training data affects the possibility of providing unifying explanations of NN learning dynamics remains an open question.
\par 
To assess the influence of training data sensitivity on the degree of unification of explanations of NN learning dynamics, studies are needed that investigate whether a given explanatory scheme, an argument pattern or explanations that contain a certain generalization, of a dynamical phenomenon is robust across distinct training data distributions and orders. A robustness target is considered robust with respect to a given robustness modifier if the interventions within the robustness domain do not cause changes in the target that exceed the specified tolerance \parencite[p. 8]{Robustness}. In studies examining the degree of unification of an explanatory scheme with respect to variations in training data distribution or order, the explanatory scheme would be treated as the robustness target, while the distributional or ordering variations of the training data would serve as robustness modifiers.
\par 
Robustness analyses are ubiquitous in NN research. However, the most common type focuses on the robustness of model performance, not on explanations. To our knowledge, no studies have analysed the robustness of \textit{explanations} of dynamical phenomena in NN learning with respect to training data distribution or order. Consequently, whether and to what extent the sensitivity of NN learning to training data affects our ability to provide unifying explanations of its dynamical behaviour remains an open question.
\par 
Nevertheless, like sensitivity on the weight initialization strategy, sensitivity to the training data can lead to a lack of pragmatic understanding, since even expert practitioners are often unable to anticipate qualitatively characteristic effects of the training distribution or the order of presentation on the learning dynamics. Consider, for example, adversarial training \parencite[]{adversarial_training}. The authors show that a visually imperceptible perturbation of a single training image can cause the prediction function to flip its predictions for 16 out of 30 test images compared to training on the unperturbed dataset. This indicates that the resulting prediction functions are distinct, implying that the two learning processes, one with and one without the perturbation, followed different weight dynamics. Yet because the perturbation is visually imperceptible, researchers cannot  anticipate visually whether the two learning dynamics will differ qualitatively \textit{before} running the training process. Adversarial training thus illustrates that in some cases it is impossible to discern qualitatively significant consequences of the training data merely by visually inspecting it.  
\begin{table}[h!]
    \centering
    \renewcommand{\arraystretch}{2}
    \begin{tabular}{P{4cm} P{5cm} P{5cm}}
        \toprule
        & \textbf{Complex systems} & \textbf{NN learning} \\
        \midrule\midrule
        
        \textit{Feedback} 
        & Reliance on incomprehensible models 
        & Challenges of statistical inference \\
        
        \textit{Sensitive dependence on initial conditions} 
        & Challenges of statistical inference 
        & Limited pragmatic understanding \\
        
        \textit{Context sensitivity} 
        & Reduced degree of unification 
        & Reduced unification?, limited pragmatic understanding \\
        
        \bottomrule
    \end{tabular}
    \caption{Challenges arising from complexity in natural complex systems and neural network learning.}
    \label{tab:Understanding}
\end{table}
\label{subsec:6.3ContrDataSens}

\section{Conclusion}
This article shed light on how complexity contributes to learning opacity. 
Table \ref{tab:Understanding} highlights our main findings. 
\par
We argued that the learning process for a neural network (NN) is essentially a complex dynamical system.
In particular, as a complex system, it displays similar characteristics as natural complex systems, such as sensitivity to initial conditions (weights), feedback between micro-states (weights) and system properties (loss gradient) and context (data) sensitivity. We then showed how each of these properties contributes to the difficulty to understand the learning process. 
\par 
Feedback impedes a theoretical analysis of the learning process and leads to a shift towards empirical machine learning (ML) research. This 
introduces the epistemological limitations inherent to statistical inference. 
Sensitivity to the weight initialization strategy affects the researcher's pragmatic understanding, 
as it becomes hard 
to qualitatively anticipate 
consequences of particular strategies.
Pragmatic understanding 
is also affected by 
sensitivity to the training data, but it remains an open question whether the latter 
also impedes 
unifying explanations of learning phenomena.
\par 
Our main conclusion is that learning opacity, understood as difficulty to understanding crucial dynamical phenomena of NN learning, is due to the structure of complexity and the epistemological challenges that arise from it.
Thus learning opacity is not a matter of lack of access or cognitive limitation. %, but due to structure and  Instead learning opacity, understood as difficulty to understanding crucial dynamical phenomena of NN learning, is due to the structure of complexity and the epistemological challenges that arise from it. %difficulty a neural network's (NN) weight dynamics during training. 
Rather, the intrinsic structure of ML  is at the heart of opacity. And we suspect that a similar analysis can be conducted for other ML learning algorithms.
 Sensitivity to weight initialization, feedback in gradient based optimization and sensitivity to the training data are fundamental to the learning process. Damping such complexity-related properties or eliminating them would fundamentally alter how ML systems learn. Consequently, some sources of opacity in ML may be irreducible. And hence the research in finding methods to reduce opacity, other than by building interpretable ML algorithms from the outset, might be a pipe dream. The conceptual bridge drawn between 
ML systems and complex natural systems, however, suggest that frameworks developed to study the latter may provide valuable approaches for addressing challenges of understanding the former, for example  determining the appropriate level of coarse-graining for analysing and simplifying ML systems might be a promising route. 
The full potential of seeing ML learning as complex system, however, remains an open field for future research.

\printbibliography
\end{document}